\documentclass[10pt,twocolumn,letterpaper]{article}

\usepackage{iccv}              

\definecolor{iccvblue}{rgb}{0.21,0.49,0.74}
\usepackage[pagebackref,breaklinks,colorlinks,allcolors=iccvblue]{hyperref}

\usepackage{multirow}
\usepackage{multicol}
\usepackage{colortbl} 
\definecolor{lightgray}{rgb}{0.9,0.9,0.9} 

\def\paperID{3025} 
\def\confName{ICCV}
\def\confYear{2025}

\title{OpenRSD: Towards Open-prompts for Object Detection in \\ Remote Sensing Images}

\author{Ziyue Huang\\
Beihang University\\
{\tt\small ziyuehuang@buaa.edu.cn}\\
\and
Yongchao Feng\\
Beihang University\\
{\tt\small fengyongchao@buaa.edu.cn}\\
\and
Shuai Yang\\
Beihang University\\
{\tt\small buaa\_yangshuai@buaa.edu.cn}\\
\and
Ziqi Liu\\
Beihang University\\
{\tt\small liuzigi2002@buaa.edu.cn}\\
\and
Qingjie Liu\\
Beihang University\\
{\tt\small qingjie.liu@buaa.edu.cn}\\
\and
Yunhong Wang\\
Beihang University\\
{\tt\small yhwang@buaa.edu.cn}
}

\begin{document}
\maketitle

\begin{abstract}
Remote sensing object detection has made significant progress, but most studies still focus on closed-set detection, limiting generalization across diverse datasets. 
Open-vocabulary object detection (OVD) provides a solution by leveraging multimodal associations between text prompts and visual features.
However, existing OVD methods for remote sensing (RS) images are constrained by small-scale datasets and fail to address the unique challenges of remote sensing interpretation, include oriented object detection and the need for both high precision and real-time performance in diverse scenarios. 
To tackle these challenges, we propose OpenRSD, a universal open-prompt RS object detection framework. OpenRSD supports multimodal prompts and integrates multi-task detection heads to balance accuracy and real-time requirements. Additionally, we design a multi-stage training pipeline to enhance the generalization of model. 
Evaluated on seven public datasets, OpenRSD demonstrates superior performance in oriented and horizontal bounding box detection, with real-time inference capabilities suitable for large-scale RS image analysis. 
Compared to YOLO-World, OpenRSD exhibits an 8.7\% higher average precision and achieves an inference speed of 20.8 FPS. 
Codes and models will be released. 

\end{abstract}


\section{Introduction}
As an important task of remote sensing (RS) image interpretation, remote sensing object detection has witnessed significant advancements in recent years \cite{orcnn, rtmdet}. 
Nonetheless, the majority of existing studies have concentrated on closed-set object detection, which aims to optimize performance on a fixed set of categories within specific datasets by refining model architectures \cite{orcnn} or loss functions \cite{kfiou}. These models frequently exhibit poor generalization capabilities across diverse datasets and are limited to detecting objects belonging to predefined categories. 
Open-vocabulary object detection (OVD) \cite{gdino, yoloworld} in natural scenes present a viable solution to this limitation by establishing multimodal associations between text prompts and visual features, thereby relaxing the limitations of predefined categories. 
Furthermore, they leverage large-scale datasets such as image-text pairs \cite{owlv2}, high-quality detection datasets \cite{glip}, and grounding data \cite{gdino} for training, enabling robust generalization across various scences.

\begin{figure}[t]
  \centering
  \includegraphics[height=6.2cm]{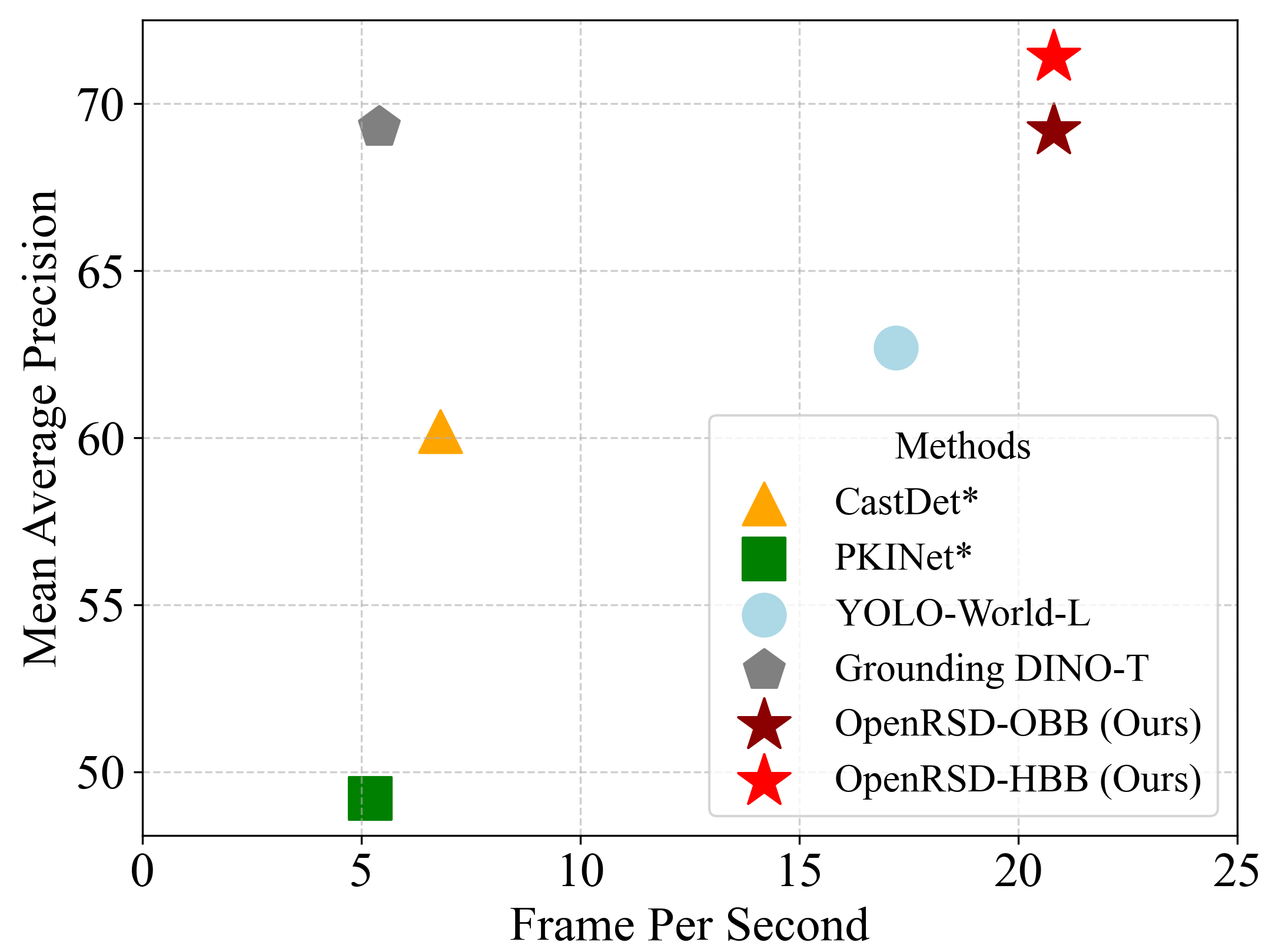}
  \caption{We compare the precision and inference speed of OpenRSD against other methods, with inference speed tested on a single 2080Ti GPU.}
  \label{fig:fig1compare}
\end{figure}

Recent studies have explored OVD models for RS images \cite{ovadetr, laedino, castdet}; however, these efforts are constrained by the use of small-scale datasets, which are inadequate to support robust generalization. 
Furthermore, they fail to address the unique requirements of RS scenarios, such as detecting small objects and precisely object orientations. 
This gap motivate us to develop a universal RS object detector with functional flexibility and strong cross-domain detection capabilities, serving as a foundational model for RS image interpretation. 
To this end, we propose a multimodal prompt-based object detector and introduce a multi-stage training pipeline to strengthen its generalization.

We improve upon the RTMDet \cite{rtmdet} framework and propose the \textbf{OpenRSD}, which achieves the best performance in both overall accuracy and inference speed, as shown in Fig. \ref{fig:fig1compare}. 
A prompt construction module is developed, utilizing pretrained models include SkyCLIP \cite{skyclip} and DINOv2 \cite{dinov2} to extract text and image prompt embeddings. 
The image prompts aim to reduce reliance on expert knowledge, which is particularly important since objects and fine-grained categories in RS images are often rare in everyday contexts. 
By incorporating random sampling during training, the detector is enabled to identify relevant object based on arbitrary prompts. 
We design two detection heads with multi-task: alignment head and fusion head. 
The alignment head performs classification by computing the similarity between detection features and prompt embeddings, offering faster inference and easy scalability for expanding the number of categories.
The fusion head introduces cross-modal interactions, achieving better detection performance. 
Meanwhile, we integrate both horizontal and oriented bounding box regression tasks into one framework. 
Both heads are trained simultaneously with minimal impact on training time, offering a flexible design to meet diverse application requirements in RS scenarios. 

To fully leverage existing datasets and enable the detector to robust detect objects in general scenarios, we propose a multi-stage training pipeline. 
Existing RS datasets are fragmentation and diversity, encompassing numerous specialized datasets, such as HRSC2016 \cite{hrsc2016} for ship and SpaceNet \cite{spacenet} for building. 
This heterogeneity poses substantial challenges for jointly trained detectors, as models trained on individual datasets often fail to generalize effectively to other scenarios, leading to severe false positives and false negatives. 
To address these challenges, we first pretrain the detector using both labeled and unlabeled data to adapt to RS data and the detection task.
Subsequently, we finetune detector and employ it to generate pseudo-labels, and leverage SkyCLIP \cite{skyclip} for filtering. 
Finally, we continue fine-tuning detector using the complete self-annotated dataset integrated with labeled data, endowing it with strong generalization performance. 

Our contributions are summarized as follows:
(1) We propose an \textbf{Open}-Prompt \textbf{R}emote \textbf{S}ensing object \textbf{D}etector (OpenRSD) that supports both oriented and horizontal object detection using either image or text prompts. 
To balance inference speed and high-precision detection, we introduce an alignment head and a fusion head, each specifically designed to address these distinct requirements.
(2) We design a multi-stage training pipeline consists of pretraining, finetuning, and self-training to enhance the detector's generalization and cross-domain detection capabilities. 
Furthermore, we construct a large-scale training dataset, ORSD+, which comprises over 470k images spanning 200 categories. 
(3) We evaluate OpenRSD on seven public datasets. 
For oriented object detection, OpenRSD demonstrates superior average performance. 
For horizontal bounding box detection, OpenRSD achieves performance comparable to the high-precision baseline model, Grounding DINO \cite{gdino}, but with significantly higher inference speed. It runs at a inference speed of 20.8 FPS, making it well-suited for large-scale RS image analysis.

\begin{figure*}[ht!]
  \centering

  \includegraphics[height=7.8cm]{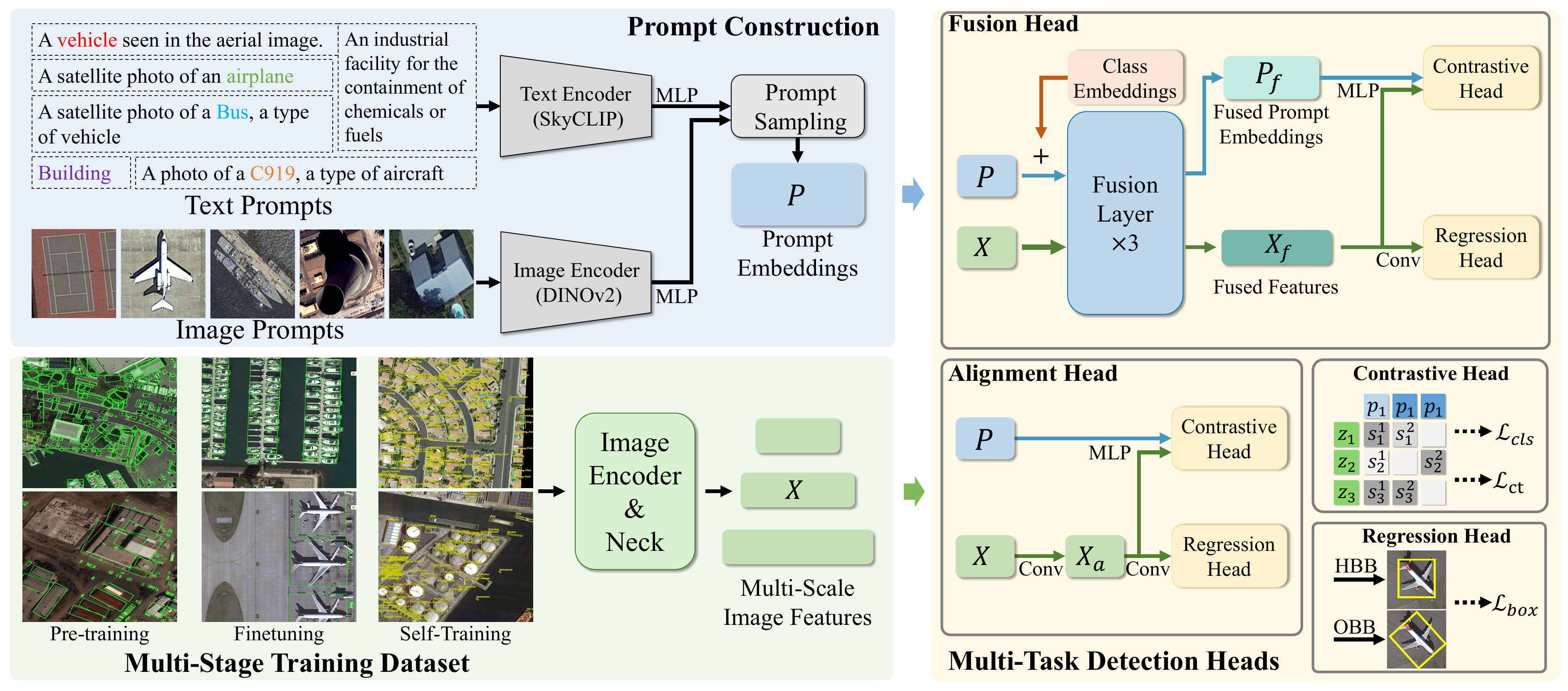}
  \caption{
  OpenRSD consists of three components: multi-scale image feature encoding, prompt construction, and multi-task detection heads.
  The prompt construction randomly samples multiple text or image prompts for each category and encodes them into prompt embeddings. 
  The multi-task detection heads leverage the correlation between prompt embeddings and image features to perform detection, seamlessly integrating fusion-based and alignment-based open-prompt structures within one framework. 
  This design enables support for a wide range of prompt-based classification and regression tasks. 
  The alignment head offers higher speed and greater vocabulary scalability, while the fusion head achieves better precision, allowing the model to adapt to different application scenarios. 
  To enhance generalization, we further employ a multi-stage training pipeline.
  }
  \label{fig:method}
\end{figure*}

\section{Related Work}
\subsection{Open-Set Object Detection}
Open-set object detection models utilize multimodal prompts \cite{gdino, trex} to detect relevant objects. 
Early works \cite{ViLD} adhered to strict OVD settings \cite{zareian2021open}, training in base classes and testing in unseen classes, using conventional datasets like COCO \cite{coco} for research. 
Recent studies have attempted to enhance generality by expanding available datasets through multitask learning \cite{gdino, detclipv3} or self-training \cite{owlv2}, and by conducting large-scale image-text alignment pre-training. 
In terms of model architecture, GLIP \cite{glip}, Grounding DINO \cite{gdino}, and OV-DINO \cite{ovdino} introduced deep cross-modal fusion, while YOLO-World incorporated lightweight feature interaction modules during multi-scale feature fusion. 
DetCLIP \cite{detclipv3} and OWL-ViT \cite{owlv2} adopted simple feature alignment combined with large-scale pre-training to improve model generalization. 
Cross-modal fusion helps enhance feature alignment and information interaction, improving detection performance, but increases the difficulty of detecting large vocabularies.
To meet the requirements of both efficiency and accuracy in RS scenarios, our proposed OpenRSD integrates both architectures into an efficient detector, enhancing the model's generality across different scenarios. 
Additionally, we introduce a multi-prompt design to reduce reliance on expert knowledge in remote sensing. 

\subsection{Object Detection in Remote Sensing}
Real RS scenarios present challenges such as extreme object scales \cite{cheng2023towards}, diverse object variations \cite{sun2022fair1m}, and arbitrary object orientations \cite{xia2018dota}. 
Existing works often focus on closed-set object detection, improving structures \cite{cai2024poly} and loss functions \cite{xu2024rethinking} but failing to detect new categories. 
Recently, some studies have extended detectors to the realm of open vocabulary.
DescReg \cite{zang2024zero} and CastDet \cite{castdet} use alignment-based approaches to achieve open-vocabulary detection, while OVA-DETR \cite{ovadetr} and LAE-DINO \cite{laedino} introduce deep cross-modal fusion. 
However, constrained by dataset scale, training methods, or model architecture, these models struggle to handle real RS scenarios. 
Our proposed OpenRSD improves upon existing state-of-the-art RS object detector \cite{rtmdet}, making it most suitable for the task. 
Furthermore, a multi-stage training pipeline is adopted to enhance the model's generalization and cross-domain capabilities.

\section{Method}

\subsection{Overview}
We propose OpenRSD, an RS object detector capable of detecting objects based on either image or text prompts. 
As illustrated in Fig. \ref{fig:method}, the model comprises an image encoder and a neck for extracting multi-scale image features $X$, a prompt construction module, and two detection heads with multi-task that operate on prompt embeddings $P$. 
Specifically, the detection heads include two parallel components: the alignment head and the fusion head, which employ feature alignment and cross-modal fusion, respectively, to effectively utilize prompts for detection. 
Additionally, we introduce a multi-stage training pipeline to significantly improve the model’s cross-dataset generalization. 

\subsection{Prompt Construction}

OpenRSD supports both text and image prompts, achieves open-prompt object detection. 
Text prompts enable the detector to perform open-vocabulary object detection, while image prompts allow the detector to perform detection based on images even in the absence of specific category names or descriptions, thereby reducing the need for expert knowledge, especially for fine-grained categories such as aircraft models. 
We utilize the text encoder from SkyCLIP \cite{skyclip}, pre-trained on large-scale RS image-text pairs, to extract text prompt embeddings, and employ DINOv2 \cite{dinov2} to extract image prompt embeddings. 
The prompt embeddings are stored offline to reduce training and inference time. 

Specifically, text prompts include category names, descriptions, or other textual information related to the categories, as illustrated in Fig. \ref{fig:method}. 
We use GPT-4 \cite{gpt4} to generate 10 to 15 text prompts to simulate potential inputs, enhancing the diversity of text prompts. 
For image prompts, we enlarge the ground truth bounding boxes in the training set by 1.25 times to include necessary visual context information, then crop and resize them to $224 \times 224$ pixels, obtaining a large number of RS object images. 
Due to the considerable variation in the spatial resolution of RS images, certain objects, such as small vehicles or ships, may appear blurred and lack distinguishable features.
To address this issue, we sample images from each dataset to construct a classification dataset, and then train a simple classifier using DINOv2 \cite{dinov2} combined with a two-layer MLP. 
For each category, we select the top 100 images with the highest classification confidence to form the final image prompt set. 
The prompt embeddings are mapped into 256-dimensional embedding space through two separate multi-layer perceptrons (MLP) for joint training.

During each iteration, we randomly select one type of prompt for training. 
We first identify the annotated categories present in the image as positive categories. 
Simultaneously, we randomly sample negative categories to enhance robustness against irrelevant prompts. 
OpenRSD supports using multiple prompts for robust detection. 
For each category, 3 to 7 prompt embeddings are randomly sampled and fed into the detection heads for further processing.

\subsection{Alignment Head}
Alignment heads are suitable for rapid detection in large vocabulary scenarios. 
It first uses a convolution layer to obtain its own image features $X_a$.
Then, we employs the decoupled head structure \cite{rtmdet} for regression and contrastive classification, where the classification branch outputs the 256-dimensional embeddings. 
Subsequently, we utilize a shared MLP to map embeddings into the same semantic space, generating output embeddings $Z = \{z_i\}_{i=1}^N \in \mathbb{R}^{N \times 256}$ and prompt embeddings $P = \{p_j\}_{j=1}^M \in \mathbb{R}^{M \times 256}$, where $N$ and $M$ denote the number of embeddings, respectively. 
Let $L = \{l_j\}_{j=1}^M$ denote the ground truth labels of the prompt embeddings. 
we compute the maximum normalized inner product between $z_i$ and the intra-class prompt embeddings as the logits for class $c$, denoted as $s_i^c$:

\begin{equation}
s_i^c = \max_{j \in \{j |l_j = c\}}\left( \alpha \frac{z_i \cdot p_j}{\|p_j\|} + \beta \right)
\end{equation}
Here, $\alpha$ and $\beta$ are adjusted during training to ensure stability \cite{yoloworld}.
Finally, the logits for each class are aggregated, and the classification loss $\mathcal{L}_{\text{cls}}$ \cite{rtmdet} is computed.

In addition to the classification loss, a contrastive loss is introduced to further stabilize the feature alignment.
Specifically, for each prompt embedding $p_j$, we select the most similar prediction embedding $z_i^*= \arg\max_{z_i} \frac{p_j \cdot z_i}{\|p_j\| \|z_i\|}$, and apply a supervised contrastive loss \cite{khosla2020supervised}:
\begin{equation}
\mathcal{L}_{\text{ct}} = \sum_i -\frac{1}{|P(i)|} \sum_{j \in P(i)} \log \frac{e^{(z_i^* \cdot p_j / \tau)}}{\sum_{k \in A(i)} e^{(z_i^* \cdot p_j / \tau)}}
\end{equation}
Here, $A(i)$ denotes the set of indices excluding $i$, $P(i)$ is the subset of $A(i)$ where the predicted and ground truth labels match, and $\tau$ is the temperature parameter, set to 0.1. 
Total loss for the alignment head is:
\begin{equation}
    \mathcal{L}_{\text{aln}} = \mathcal{L}_{\text{ct}} + \mathcal{L}_{\text{cls}} + \mathcal{L}_{\text{box}}
\end{equation}
where $\mathcal{L}_{\text{box}}$ is the regression loss \cite{rtmdet}, which includes both horizontal and oriented box regression.

\subsection{Fusion Head}

The fusion head introduces a lightweight cross-modal fusion to facilitate interaction between prompt embeddings and detector features, enabling improved detection performance.
Additionally, we introduce a class embedding to explicitly differentiate between distinct categories. 
Specifically, for each image, we gather the corresponding annotated categories and assign a random category ID. 
This ID is then mapped to a learnable class embedding, which is added to all prompt embeddings of that class. 
The class embedding enables the fusion head to dynamically manage detection tasks at varying granularities, preventing conflicts between tasks. 
The fusion module comprises three layers, each defined as follows: 

\begin{align}
P' &= \text{LN}(\text{MHCA}(P_i, X_i) + P) \\ 
P_{i+1} &= \text{LN}(\text{MLP}(P') + P') \\
X' &= \text{LN}(\text{MHCA}(X_i, P_{i+1}) + X_i) \\
X_{i+1} &= \text{LN}(\text{MLP}(X') + X')
\end{align}
where MHCA is multi-head cross-attention, LN is layer normalization, and MLP is multi-layer perceptron. 
The classification, supervised contrastive, and regression losses are computed same to the alignment head, using the fused features $X_f$ and fused prompt embeddings $P_f$, with their sum denoted as $\mathcal{L}_{\text{fus}}$. The overall loss for OpenRSD is:
\begin{equation}
   \mathcal{L}_{\text{det}} = \mathcal{L}_{\text{fus}} + \mathcal{L}_{\text{aln}}
\end{equation}

\begin{figure*}[t]
  \centering
  \includegraphics[height=3.0cm]{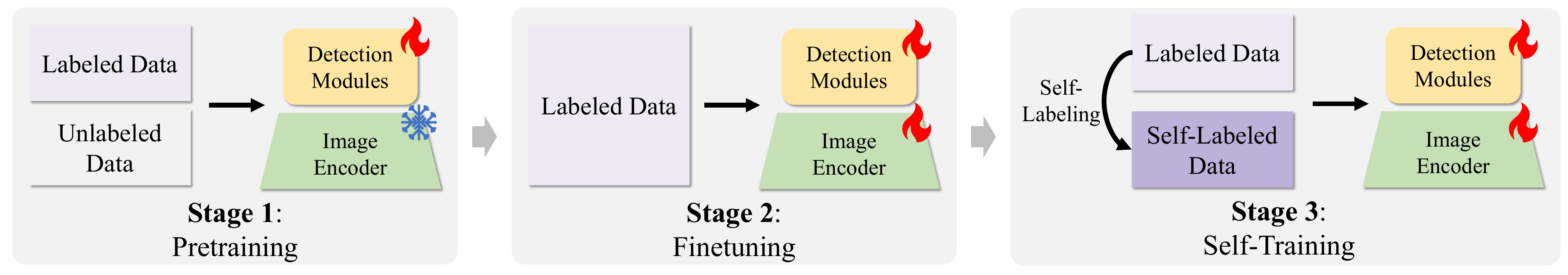}
  \caption{The multi-stage training pipeline includes pretraining, fine-tuning, and self-training. Pretraining trains only the detection modules to adapt to RS detection task. Fine-tuning stage enables the detector to detect arbitrary objects in RS images. Self-training enhances cross-scenario generalization.}
  \label{fig:training}
\end{figure*}

\subsection{Multi-Stage Training}
To ensure the detector to handle diverse prompt inputs and varying RS scenarios, we introduce a multi-stage training pipeline, as illustrated in Fig. \ref{fig:training}. 

In the first stage, we pretrain only the detection modules with a combination of labeled and unlabeled datasets, enabling the detector to adapt to the RS object detection task. 
For labeled datasets, we select seven RS datasets that cover common categories. 
We use the oriented bounding box annotations to generate minimum enclosing horizontal boxes, which are then employed for training the horizontal box regression. 
The Million-AID dataset \cite{maid} is employed as the unlabeled dataset.
Specifically, we use SAM \cite{sam} to generates region proposals, which are subsequently cropped and processed using DINOv2 \cite{maid} to extract object embeddings \cite{huang2025mutdet}. 
These embeddings are then clustered, and the average embeddings of each cluster is utilized as prompt embeddings during training. 

In the second stage, we collect additional data, resulting in a total of ten datasets at various levels of category granularity. 
During training, an empirical sampling rate is employed to sample the datasets. 
Despite achieving exceptional performance, we observe that the detector often failed to accurately detect objects like ships and buildings and can not generalize to mixed prompts from multiple datasets. 
These issues primarily because RS datasets often restrict to specific categories, such HRSC2016 \cite{hrsc2016} and SpaceNet \cite{spacenet} dataset, resulting in significant domain gaps that limit the model’s generalization. 
Moreover, the absence of training on mixed prompt scenarios leads to inconsistencies in real-world applications.
To tackle these challenges, we implement a straightforward self-training strategy.

In the third stage, we use the fine-tuned detector to complete the annotations of the labeled dataset, aiming to comprehensively cover all objects within the predefined 200 categories. 
Upon completion, each image is assigned an independent category list to enhance the semantic diversity of prompts. 
First, we perform detection using prompt embeddings constructed from each dataset, setting a score threshold of 0.3 to select high-confidence detection results, and only retain detection boxes un-overlap with the ground truth. 
For detections with high overlap with ground truth, we check whether the predicted category and the ground truth category belong to the same parent class by a pre-defined category tree \cite{maid}. 
Predicted categories with different parent classes from the ground truth are recorded as hard negative categories and added to the category list. 
Next, we merge all predictions, and apply class-agnostic NMS to remove duplicate detection results. 
For objects with pixels larger than $16\times16$, we use SkyCLIP \cite{skyclip} for further filtering. 
Specifically, we extract the textual features of predicted categories and the visual features of boxes using SkyCLIP's text encoder and image encoder, respectively, and calculate their cosine similarity. 
Based on the filtering visualization results, we remove predictions with a similarity score below 0.24. 
The retained predicted categories are added to the category list of the corresponding image. 
Finally, we continue training the finetuned model using a combination of the self-labeled dataset and the original labeled dataset. 
The self-labeled dataset is sampled at half the rate of the original labeled dataset to preserve original capabilities while improving cross-domain performance.

\begin{table*}[htbp]
  \centering
  \caption{Comparison on seven public datasets. The evaluation includes both oriented bounding box (OBB) and horizontal bounding box (HBB) detection, using AP50 as the evaluation metric. Models marked with * are trained using the proposed ORSD+ dataset. Inference speed, measured in FPS, is evaluated on a single 2080Ti.}
  \footnotesize
    \setlength{\tabcolsep}{3.2pt} 
    \begin{tabular}{c|c|ccccccc|cc}
    \hline
    & Backbone & DIOR-R & DOTA-v1.0 & DOTA-v2.0 & FAIR-1M-2.0 & WHU-Mix & SpaceNet & HRSC2016 & Average & FPS \\
    \hline
    \textit{OBB Evaluation} &     &     &     &     &     &     &     &     &     &  \\
    RetinaNet-OBB \cite{ross2017focal}\textcolor{gray}{\tiny [CVPR17]} & ResNet-50 & 51.7  & 61.3  & 60.0  & 35.9  & 77.3  & 43.2 &41.2 &	52.9   & 20.6 \\
    RoI Transformer \cite{roitrans}\textcolor{gray}{\tiny [CVPR19]} & ResNet-50 & 61.6  & 73.1  & 64.1  & 43.5  & 79.6  & 45.7  & 51.9 &	59.9   &  11.2\\
    S$^2$ANet \cite{s2anet}\textcolor{gray}{\tiny [TGRS21]} & ResNet-50 & 61.3  & 70.8  & 63.8  & 44.0  & 79.5  & 45.5  &64.2&	61.3   & 16.9\\
    R$^3$Det-KFIoU \cite{kfiou}\textcolor{gray}{\tiny [ArXiv22]} & ResNet-50 & 57.2  & 66.6  & 61.1  & 41.4  & 78.5  & 44.1  & 50.8 & 57.1   &  18.2 \\
    Oriented R-CNN \cite{orcnn}\textcolor{gray}{\tiny [ICCV21]} & ResNet-50 & 59.7  & 70.2  & 63.0  & 43.2  & 79.3  & 44.9  &57.2 &	59.6  &  7.0\\
    Oriented R-CNN \cite{orcnn}\textcolor{gray}{\tiny [ICCV21]} & Swin-T & 65.8  & 76.0  & 67.4  & 45.0  & 79.3  & 44.6  & 79.1 &	65.3   & 6.9\\
    MTP \cite{wang2024mtp}\textcolor{gray}{\tiny [JSTARS24]} & ViT-B & 72.1  & 76.8 & 67.8  & 44.2  & \textbf{80.0}  & 46.7  &80.1  &	66.8  & 4.7\\
    RTMDet-R-M \cite{rtmdet}\textcolor{gray}{\tiny [ArXiv22]} & RTMDet-M & 69.4  & 74.5  & 61.4  & 46.9  & 79.9  & 47.3  & 79.5  & 65.6  & \textbf{35.7}  \\
    RTMDet-R-L \cite{rtmdet}\textcolor{gray}{\tiny [ArXiv22]}  & RTMDet-L & 70.1  & 71.5  & 66.7  & \textbf{47.6}  & 79.9  & 47.4  & 79.9  & 66.2  & 23.8 \\
    CastDet$^*$ \cite{castdet}\textcolor{gray}{\tiny [ECCV24]}  & ResNet-50  & 63.3  & 72.1  & 63.1  & 37.7  & 77.7  & 45.0  & 74.1  & 61.9  & 6.8\\
    PKINet$^*$ \cite{pkinet}\textcolor{gray}{\tiny [CVPR24]}  & PKINet-T & 56.7  & 64.9  & 59.6  & 34.6  & 62.6  & 34.6  & 41.3  & 50.6  & 5.2 
    \\
    \rowcolor{lightgray}
    OpenRSD$^*$ (Text Prompt) & RTMDet-L & \textbf{73.7}  & \textbf{76.9}  & \textbf{70.1}  & 46.1  & 79.7  & \textbf{50.2}  & 88.1  & \textbf{69.3}  & 20.8 \\
    \rowcolor{lightgray}
    OpenRSD$^*$ (Image Prompt) & RTMDet-L & \textbf{73.7}  & 76.7  & 67.8  & 44.1  & 79.6  & \textbf{50.2}  & \textbf{88.6}  & 68.7  & 20.8\\

    \hline
    \textit{HBB Evaluation} &     &     &     &     &     &     &     &     &     &  \\
    YOLO-World-L \cite{yoloworld}\textcolor{gray}{\tiny [CVPR24]}  & YOLOv8-L  & 73.2  & 70.8  & 58.0  & 41.1  & 82.6  & 53.7  & 59.4  & 62.7 & 17.2\\
    Grounding DINO-T \cite{gdino}\textcolor{gray}{\tiny [ECCV24]}  & Swin-T & \textbf{78.7}  & \textbf{78.7}  & \textbf{71.8}  & \textbf{46.0}  & \textbf{85.4}  & \textbf{56.5}  & 68.0  & 69.3 & 5.4\\
    \rowcolor{lightgray}
    
    OpenRSD$^*$ (Text Prompt) & RTMDet-L& 76.7  & 77.7  & \textbf{71.8}  & 45.8  & 85.0  & 55.4  & \textbf{87.7}  & \textbf{71.4} &  \textbf{20.8} \\
    \rowcolor{lightgray}

     OpenRSD$^*$ (Image Prompt) & RTMDet-L& 76.7  & 77.3  & 69.8  & 43.9  & 84.7  & 55.4  & 87.3  & 70.7 &  \textbf{20.8}\\
    \bottomrule
    
    \hline
    \end{tabular}%
  \label{tab:Table1_General}%
    
\end{table*}

\begin{table}[t]
  \centering
  \caption{The datasets used in the multi-stage training pipeline. ORSD-Pre dataset is used for pretraining, and ORSD+ dataset is used for finetuning and self-training.}
      \scriptsize
        \setlength{\tabcolsep}{1.4pt} 
    \begin{tabular}{c|c}
            \hline

    Name & Datasets  \\
            \hline
    \multirow{3}[0]{*}{ORSD-Pre} & DOTA-v2.0 \cite{dota2}, DIOR-R \cite{diorr}, FAIR1M-2.0 \cite{sun2022fair1m},  SpaceNet \cite{spacenet}\\
        &  Xview \cite{xview}, HRSC2016 \cite{hrsc2016}, GLH-Bridge \cite{li2024learning}, Million-AID \cite{maid}\\
        \hline
        
    \multirow{4}[0]{*}{ORSD+} 
    & DOTA-v2.0 \cite{dota2}, DIOR-R \cite{diorr}, FAIR1M-2.0 \cite{sun2022fair1m}, SpaceNet \cite{spacenet}\\
        & SpaceNet \cite{spacenet}, Xview \cite{xview}, HRSC2016 \cite{hrsc2016}, GLH-Bridge \cite{li2024learning} \\
        &  fMoW \cite{christie2018functional}, WHU-Mix \cite{whumix}, ShipRSImageNet \cite{zhang2021shiprsimagenet}\\
        \hline

    \end{tabular}%
    \vspace{-2em}

  \label{tab:Table10_Dataset}%
\end{table}%

\section{Experiments}

\subsection{Datasets}

We use 11 datasets for training and evaluation, which can be categorized into three major groups:

\textbf{General Datasets.}
DIOR-R \cite{diorr} dataset contains 20 categories. The training and validation sets consist of 11,725 images, each with a size of 800 $\times$ 800 pixels. We use the training and validation sets for training and evaluate on the test set. 
DOTA-v1.0 \cite{xia2018dota} and DOTA-v2.0 \cite{dota2} dataset contain 15 and 18 categories, respectively, include diverse RS images with varying scenes.  
For DOTA-v1.0, we clip images into 800 $\times$ 800 pixels with a stride of 600 \cite{kfiou}. 
For DOTA-v2.0, we clip images into 1,024 $\times$ 1,024 pixels with an overlap of 500 and apply multi-scale augmentation. 
We train on the training set and evaluate on the validation set.

\textbf{Specific Datasets.}
WHU-Mix \cite{whumix} is a large-scale building detection dataset containing 43,778 images with spatial resolutions ranging from 0.09 to 2.5 meters. 
We train on the training and validation sets and evaluate on the test set. 
SpaceNet \cite{spacenet} is a building detection dataset covering multiple cities. 
We merge four cities to construct a training set of 10,152 images and evaluate on the merged test set. 
GLH-Bridge \cite{li2024learning} is a bridge dataset, in which images are clipped into 1,024 $\times$ 1,024 pixels with an overlap of 200.

\textbf{Fine-Grained Datasets}
FAIR1M-2.0 \cite{sun2022fair1m} dataset includes 37 fine-grained categories for aircraft, ships, scenes, and vehicles. 
We clip images into 800 $\times$ 800 pixel patches with a stride of 400, using validation set for evaluation. 
Xview \cite{xview} dataset involves 60 fine-grained categories. 
We randomly select 700 images as the training set and 146 images for testing, clip images into 800 $\times$ 800 pixels with an overlap of 200, HRSC2016 \cite{hrsc2016} and ShipRSImageNet \cite{zhang2021shiprsimagenet} are fine-grained ship detection datasets. 
We train on the training and validation sets and evaluate on the test set. 
FMoW \cite{christie2018functional} dataset contains 62 foreground classes, covering common scenes and fine-grained categories of buildings. 

Data synthesis has become a crucial source for training large models \cite{wang2024survey}. 
To this end, we employ multi-stage training pipeline to generate more data.
The details of the dataset used for pipeline are shown in Tab. \ref{tab:Table10_Dataset}. 
Among them, ORSD-Pre is used for pretraining, while ORSD+ serves as the fine-tuning dataset. 
ORSD+ contains a total of 474,058 images across 200 categories.

\subsection{Experimental Setup}

\textbf{OpenRSD.}
We use DINOv2-ViT-L \cite{dinov2} to extract image prompt embeddings and the text encoder of SkyCLIP-ViT-L \cite{skyclip} to extract text prompt embeddings. 
These embeddings are extracted offline and stored as fixed prompt dictionaries. 
We implement OpenRSD based on the RTMDet-L \cite{rtmdet} in MMRotate \cite{zhou2022mmrotate} framework.
The model is trained using the Adam optimizer \cite{loshchilov2018decoupled} with an initial learning rate of 2.5 $ \times 10^{-4}$, 
a weight decay of 5 $\times 10^{-2}$, and a cosine annealing scheduler that reduces the learning rate to 5\% of its initial value mid-training. 
Training is performed on four 3090Ti GPUs with a batch size of 16. 
Pretraining involves 360k iterations with input of 896 $\times$ 896 pixels, while fine-tuning and self-training each consist of 288k iterations on images with input of 832 $\times$ 832 pixels.


\textbf{Compared Methods.}
We compare our method with SOTA oriented object detection methods, including RetinaNet-OBB \cite{ross2017focal}, RoI-Transformer \cite{roitrans}, S$^2$ANet \cite{s2anet},  R$^3$Det-KFIoU \cite{kfiou}, Oriented R-CNN \cite{orcnn}, MTP \cite{wang2024mtp}, and RTMDet \cite{rtmdet}. 
These models are trained and tested on their respective datasets. 
To ensure a fair comparison, we also include OVD methods that could jointly trained using ORSD+, namely CastDet$^*$ \cite{castdet} and PKINet$^*$ \cite{pkinet}. 
We also compare with two powerful OVD methods, Grounding-DINO \cite{gdino} and YOLO-World \cite{yoloworld}, in horizontal bounding box detection. 
Since these two detectors have been sufficiently pre-trained, fine-tuning them on each dataset can demonstrate their optimal performance \cite{laedino}. 

The training schedule is adjusted based on the model architectures and datasets: the default schedule consists of 12 epochs, while single-stage detectors, including RetinaNet-OBB, RTMDet, S$^2$ANet, and YOLO-World, follow a 36-epoch schedule. 
For the HRSC2016 dataset, the schedule is tripled. 
Both CastDet$^*$ and PKINet$^*$, same as OpenRSD, are trained for 288k iterations. 
The average precision at IoU 0.5 (AP50) in DOTA \cite{xia2018dota} is used as the evaluation metric.

\subsection{Main Results}

Tab. \ref{tab:Table1_General} compares the performance of various models across seven datasets. 
For OpenRSD, we report the results using text and image prompts separately, with the number of prompts set to 7.
For oriented object detection (OBB), OpenRSD achieves the best average performance, obtaining either the best or second-best precision across most datasets while maintaining a real-time inference speed of 20.8 FPS. 
Notably, on widely-used datasets such as DIOR-R, DOTA-v1.0, and DOTA-v2.0, OpenRSD outperforms the baseline RTMDet-R-L by margins of 3.5\%, 5.2\%, and 3.5\%, respectively. 
OpenRSD also outperform the MTP, which relies on ViT-B \cite{vit} large-scale pretraining, across most datasets, while functioning as a prompt-based detector that requires no fine-tuning. 
It demonstrates significant improvements on the HRSC2016 dataset, excelling in both coarse-grained and fine-grained RS object detection tasks. 
While CastDet$^*$, which also employs multi-dataset joint training,  showing impo compared to models fine-tuned on individual datasets, its framework design limitations resulted in inferior performance compared to OpenRSD. 
CastDet$^*$, which also employs multi-dataset joint training, achieves limit improvement compared to baseline Oriente R-CNN, due to structural limitations and training procedures. 

For horizontal box detection (HBB), OpenRSD exhibit a substantial accuracy advantage over YOLO-World-L, which also built on the YOLO architecture. 
Specifically, OpenRSD outperform YOLO-World-L by 4.0\%, 6.5\%, 13.8\%, and 4.6\% on DIOR-R, DOTA-v1.0, DOTA-v2.0, and FAIR1M-2.0, respectively, while achieving comparable inference speed. 
The performance of OpenRSD on most datasets is on par with the high-precision Grounding-DINO-T method. 
Notably, OpenRSD is four times faster than Grounding-DINO-T, making it more suitable for large-scale RS image analysis.

\begin{table}[t]
  \centering
  \caption{Evaluation of cross-dataset performance. All models, except OpenRSD, are fully trained on DOTA-v2.0. 
  Models marked with * are trained using the proposed ORSD+ dataset.
  }
  \footnotesize
    \setlength{\tabcolsep}{4.0pt} 
    \begin{tabular}{c|cccc|c}
    \hline
       Methods & VEDAI & CORS & DOSR & SODA-A & Average \\
    \hline
    RetinaNet-OBB \cite{ross2017focal} & 60.9  & 71.6  & 76.3  & 54.5  & 65.8  \\
    RoI Transformer \cite{roitrans} & 62.2  & 72.2  & 90.1  & 63.0  & 71.9  \\
    S$^2$ANet \cite{s2anet} & 62.2  & 78.5  & 89.7  & 62.5  & 73.2  \\
    Oriented R-CNN \cite{s2anet} & 62.1  & 72.2  & 90.1  & 62.9  & 71.8  \\
    RTMDet-R-M \cite{rtmdet} & 66.8  & 78.4  & 90.4  & 62.9  & 74.6  \\
    RTMDet-R-L \cite{rtmdet} & 67.8  & 78.6  & 89.9  & 67.1  & 75.9  \\
    CastDet$^*$ \cite{castdet} & 66.3  & 72.1  & 90.2  & 60.1  & 72.2 \\
   PKINet$^*$ \cite{pkinet} & 64.1  & 72.1  & 90.3  & 60.3  & 71.7 \\
            \rowcolor{lightgray}
    OpenRSD & \textbf{69.7}  & \textbf{79.8}  & \textbf{90.6}  & \textbf{72.3}  & \textbf{78.1}  \\
    \bottomrule
    \end{tabular}%
  \label{tab:Table3_CrossDataDOTA2}%
    
\end{table}
\begin{table}[t]
  \centering
  \caption{Ablation study on model components.}
    \footnotesize
  \setlength{\tabcolsep}{3.0pt} 
    \begin{tabular}{ccc|ccc}
    \hline
    Alignment & Fusion & Class & \multirow{2}[2]{*}{DIOR-R} & \multirow{2}[2]{*}{DOTA-v2.0} & \multirow{2}[2]{*}{FAIR1M-2.0} \\
    Head & Head & Embeddings &     &     &  \\
    \hline
    \checkmark   &     &     & 71.5 & 68.8 & 45.4 \\
    \checkmark   & \checkmark   &  &  \textbf{73.1}   &   69.5  & 45.3     \\
    \checkmark   &\checkmark   & \checkmark   & 72.9 & \textbf{70.4} & \textbf{45.9} \\
    \hline
        &     &     & SpaceNet & HRSC2016 & Average \\
    \hline
    \checkmark   &     &     & \textbf{47.8} & 84.3 & 63.3 \\
    \checkmark   & \checkmark   &   & \textbf{47.8}  & 73.1    & 61.8     \\
    \checkmark   & \checkmark   & \checkmark   & 47.5 & \textbf{84.6} & \textbf{64.3} \\
    \hline
    \end{tabular}%
  \label{tab:Table4_AblationModules}%
    
\end{table}
\begin{table}[t]
  \centering
  \caption{Detection performance with different types of head.}
  \footnotesize
    \setlength{\tabcolsep}{5pt} 
    \begin{tabular}{c|cccc}
    \hline
    Head Type & DIOR-R & DOTA-v1.0 & DOTA-v2.0 & FAIR1M-2.0 \\
    \hline
    Alignment & 71.4  & 76.4  & 68.2  & 45.2  \\
    Fusion & \textbf{72.9}  & \textbf{76.6}  & \textbf{70.4}  & \textbf{45.9}  \\
    \hline
        & WHU-Mix & SpaceNet & HRSC2016 & Average \\
    \hline
    Alignment & \textbf{79.7}  & \textbf{47.8}  & 80.5  & 67.0  \\
    Fusion & 79.4  & 47.5  & \textbf{84.6}  & \textbf{68.2}  \\
    \hline
    \end{tabular}%
  \label{tab:Table6_Diff_Eval_Methods}%
\end{table}

Tab. \ref{tab:Table3_CrossDataDOTA2} evaluates the cross-domain performances of various models. 
We select four datasets, i.e. VEDAI \cite{vedai}, CORS \cite{cors}, DOSR \cite{dosr}, and SODA-A \cite{duan2022soda}. 
These datasets are excluded from the training process, as test sets and standardized their categories to align with the categories in DOTA-v2.0 \cite{dota2}. 
The compared methods are multi-scale trained on DOTA-v2.0, ensuring their generalization on common RS object detection. 
As shown in Tab. \ref{tab:Table3_CrossDataDOTA2}, RTMDet demonstrates best average performance, highlighting the importance of model structure. 
Compared to RTMDet-L baseline, OpenRSD achieves an average improvement of 2.2\%, with particularly notable gains of 5.2\% in the small object detection dataset SODA-A \cite{duan2022soda}. 

\subsection{Ablation Studies}

We conduct a comprehensive analysis of the modules and training strategy of OpenRSD. 

\textbf{Modules.}
Tab. \ref{tab:Table4_AblationModules} examines the impact of individual modules. 
Introducing the alignment head alone achieves near-optimal performance in the building detection (SpaceNet) while remaining compatible with fine-grained detection (FAIR1M-2.0 and HRSC2016). 
The fusion head improves performance on DIOR-R and DOTA-v2.0 by 1.6\% and 0.7\%, respectively, though at the cost of reduced fine-grained recognition capability. 
By incorporating class embeddings, OpenRSD achieves greater training stability and better performance across almost all dataset. 
Tab. \ref{tab:Table6_Diff_Eval_Methods} compares detection performance when using different head types for inference after joint training. 
The fusion head outperform the alignment head by an average margin of 1.2\%, demonstrating that deep interactions between image features and prompt embeddings enhance the model's generalization capability from multi-dataset joint training.

\begin{table}[t]
  \centering
  \caption{Comparison of performance under different prompt types and quantities.}
    \footnotesize
      \setlength{\tabcolsep}{10pt} 
    \begin{tabular}{c|ccccc}
    \hline
    \multirow{2}[1]{*}{Prompt Type} & \multicolumn{5}{c}{\#Number of Prompts} \\
        & 1   & 3   & 5   & 10  & 20 \\
    \hline
    \textit{\textbf{DOTA-v2.0}} &     &     &     &     &  \\
    Image Prompt & 65.7  & 67.8  & 67.8  & 68.2  & 68.0  \\
    Text Prompt & 70.1  & 70.2  & 70.1  & 70.1  & 70.1  \\
    \hline
    \textit{\textbf{HRSC2016}} &     &     &     &     &  \\
    Image Prompt & 87.9  & 87.7  & 88.1  & 88.1  & 88.0  \\
    Text Prompt & 88.1  & 88.3  & 88.6  & 88.0  & 87.7  \\
    \hline
        \textit{\textbf{FAIR1M-2.0}} &     &     &     &     &  \\
    Image Prompt & 44.0  & 44.0  & 44.1  & 44.1  & 44.0  \\
    Text Prompt & 46.0  & 45.9  & 46.1  & 46.1  & 46.0  \\
\hline
    \end{tabular}%
  \label{tab:Table7_Prompt}%

\end{table}
\begin{figure*}[t]
  \centering
  \includegraphics[height=7.4cm]{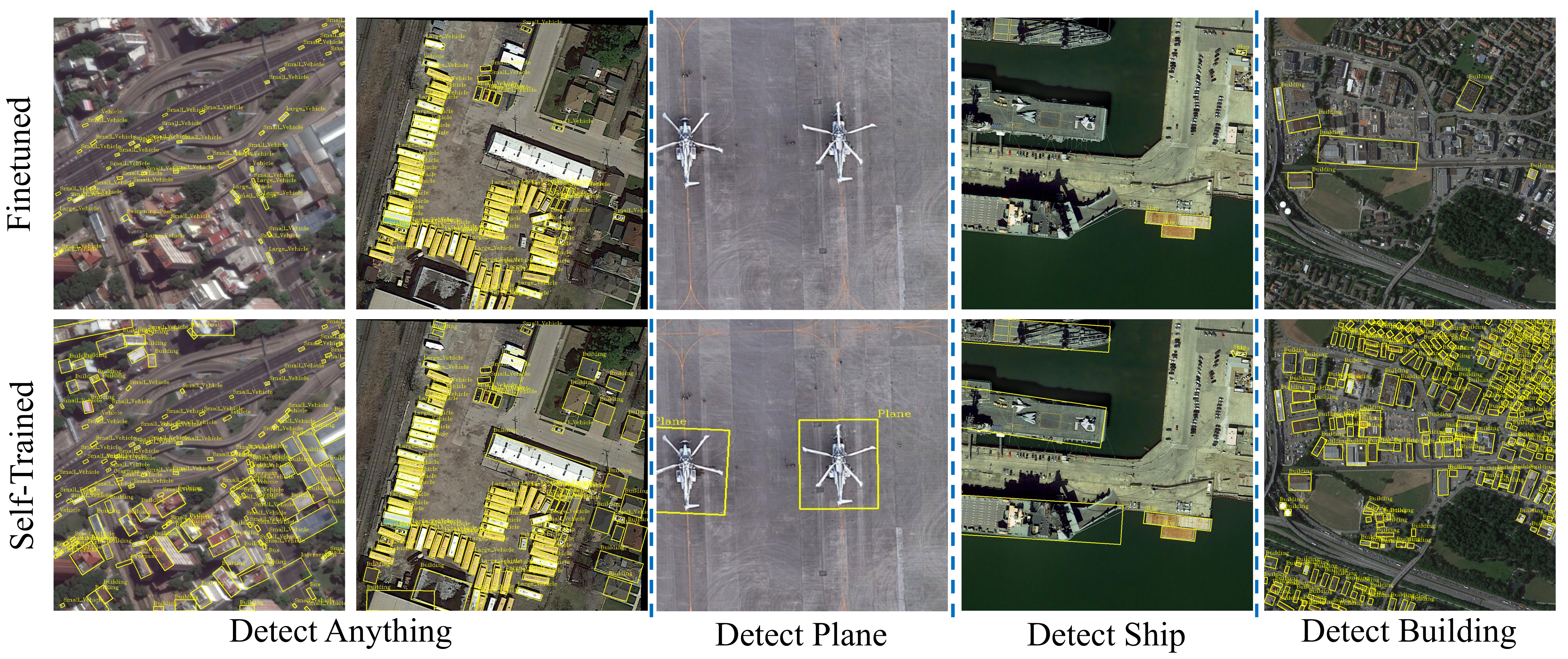}
  \caption{Visualization results on the DOTA-v2.0 \cite{dota2} validation set before (the top row) and after self-training (the bottom row), demonstrating four prompts: detecting any objects, detecting plane, detecting ship, and detecting buildings. }
  \label{fig:selftraining}
\end{figure*}

\begin{table*}[t!]
  \centering
  \caption{Ablation study of the multi-stage training strategy. For self-training, we compare self-labeling with the labeled dataset, followed by freezing the backbone or full-parameter fine-tuning. We also test self-labeling with the Million-AID \cite{maid} dataset and full-parameter fine-tuning.}
  \small
  \setlength{\tabcolsep}{6pt} 
    \begin{tabular}{ccc|cccccc}
    \hline
    Pre-training & Fine-tuning & Self-training & DIOR-R & DOTA-v1.0 & FAIR-1M-2.0 & SpaceNet & HRSC2016 & Average \\
    \hline
    \checkmark   &     &     & 69.2  & 69.2  & 21.5  & 43.3  & 33.0  & 47.2  \\
        & \checkmark   &     & 70.8  & 76.3  & 44.6  & 47.7  & 80.8  & 64.0  \\
    \checkmark   & \checkmark   &     & 72.9  & 76.6  & 45.9  & 47.5  & 84.6  & 65.5  \\
    \checkmark   & \checkmark   & Freezing Backbone & 73.7  & 75.6  & \textbf{46.0}  & 48.0  & 85.0  & 65.7  \\
    \checkmark   & \checkmark   & Full-Parameters & \textbf{73.6}  & \textbf{76.7}  & \textbf{46.0}  & \textbf{50.3}  & \textbf{87.8}  & \textbf{66.9}  \\
    \checkmark   & \checkmark   & Using Million-AID & 73.2  & 76.0  & 45.7  & 47.8  & 83.6  & 65.3  \\
    \hline
    \end{tabular}%
  \label{tab:Table5_MultiStageTraining}%
    
\end{table*}

\textbf{Prompts.}
Tab. \ref{tab:Table7_Prompt} explores the impact of different prompt types and quantities. 
Unlike few-shot fine-tuning, all parameters remain frozen, with only the number of prompts being adjusted; thus, the improvements are not significant.
Compared to image prompts, which require additional semantic extraction, text prompts inherently carry semantic information. 
It exhibit stronger generalization and greater stability when only use one prompt.
Image prompts, on the other hand, mitigate the randomness as the number increases, leading to a gradual performance improvement.


Tab. \ref{tab:Table5_MultiStageTraining} analyzes the impact of multi-stage training pipeline. 
During the pretraining stage, we combine labeled and unlabeled data to pretrain the detection module, enabling it to adapt to RS object detection. 
Although the pre-trained detector exhibits poor detection performance due to the frozen backbone and the presence of noise in the unlabeled data, it is beneficial for enhancing the final performance. 
Comparing the results in row 2 and 3 of Tab. \ref{tab:Table5_MultiStageTraining}, pretraining provides an average improvement of 1.5\%, particularly enhancing semantic discrimination capabilities, with performance gains of 1.9\%, 1.3\%, and 3.8\% on DIOR-R, FAIR1M-2.0, and HRSC2016, respectively.
Subsequent self-training using self-labeled dataset further boost detection performance. 
Experiments reveal that unfreezing the backbone for full-parameter fine-tuning led to more significant improvements, with an overall gain of 1.4\% compared to the stage before self-training.
Additionally, we test whether incorporating an external unlabeled dataset for self-training leads to further improvements. 
Specifically, we selecte 250k images from the Million-AID \cite{maid} with resolutions exceeding 512$ \times $ 512 pixels for self-training, as shown in the last row of the Tab. \ref{tab:Table5_MultiStageTraining}.
Results indicate that more data dose not lead to further performance gains, as the scale of the existing labeled datasets was already sufficient for achieving robust RS object detection. 

Fig. \ref{fig:selftraining} compares the visualization results before and after self-training for four different prompts: detecting any objects, detecting planes, detecting ships, and detecting buildings.
To achieve detection of any objects, we use the alignment head to identify all potential categories, and then use the fusion head for further detection. 
After self-training, the model exhibits fewer false positives, higher recall, and more precise localization. 
Moreover, self-training effectively improves the detection performance of planes, ships, and buildings across different scenarios.

\section{Conclusion}

In this paper, we propose OpenRSD, an open-prompt RS object detection method. 
By incorporating multi-task detection heads and a multi-stage training pipeline, OpenRSD achieves robust generalization across diverse datasets and scenarios. 
The integration of alignment and fusion heads allows for flexible and efficient detection, balancing speed and precision. 
Extensive evaluations on seven public datasets demonstrate that OpenRSD outperforms state-of-the-art models in both oriented and horizontal detection, while maintaining real-time inference.

\small
\bibliographystyle{ieeenat_fullname}
\bibliography{main}

\end{document}